\pdfoutput=1

\documentclass[11pt]{article}

\usepackage{emnlp2021}

\usepackage{times}
\usepackage{latexsym}
\usepackage{graphicx}
\usepackage[T1]{fontenc}

\usepackage[utf8]{inputenc}
\usepackage{amsmath}
\usepackage{microtype}
\usepackage{tipa}
\usepackage{xcolor}
\usepackage[normalem]{ulem}

\title{Predicting non-native speech perception using the Perceptual Assimilation Model and state-of-the-art acoustic models}

\author{Juliette Millet \\
  CoML, ENS/CNRS/EHESS/INRIA/PSL\\ 
  LLF, University of Paris, CNRS\\
  CRI, FAN, IIFR, University of Paris \\
  Paris, France\\
  \And
  {\bf Ioana Chitoran} \\
 Department of Linguistics,\\ University of Paris \\
 Paris, France\\

  \AND
  { \bf Ewan Dunbar} \\
  University of Toronto, Toronto, Canada \\
    CoML, ENS/CNRS/EHESS/INRIA/PSL, Paris, France\\
  \\
    \texttt{juliette.millet@cri-paris.org}\\
     \texttt{ioana.chitoran@univ-paris-diderot.fr } \\ 
  \texttt{ewan.dunbar@utoronto.ca}
  }
  
\begin{document}
\maketitle
\begin{abstract}
Our native language influences the way we perceive speech sounds, affecting our ability to discriminate non-native sounds. We compare two ideas about the influence of the native language on speech perception: the Perceptual Assimilation Model, which appeals to a mental classification of sounds into native phoneme categories, versus the idea that rich, fine-grained phonetic representations tuned to the statistics of the native language, are sufficient. We operationalize this idea using representations from two state-of-the-art speech models, a Dirichlet process Gaussian mixture model and the more recent wav2vec 2.0 model. We present a new, open dataset of French- and English-speaking participants' speech perception behaviour for 61 vowel sounds from six languages. We show that phoneme assimilation is a better predictor than fine-grained phonetic modelling, both for the discrimination behaviour as a whole, and for predicting differences in discriminability associated with differences in native language background. We also show that wav2vec 2.0, while not good at capturing the effects of native language on speech perception, is complementary to information about native phoneme assimilation, and provides a good model of low-level phonetic representations, supporting the idea that both categorical and fine-grained perception are used during speech perception.

\end{abstract}

\section{Introduction}

Our native language has an impact on our speech perception.
Importantly, it influences the way we perceive unfamiliar sounds---for example, from other languages.
For instance, Japanese native speakers have difficulties in telling apart the English sounds [r] and [l] \cite{yamada1992perception}, which are not in their native phoneme inventory, and confuse the English words `right' and `light' \cite{yamada1990perception}, while this contrast is highly distinct for English native speakers.

The influence of  native language on  speech perception is not yet fully understood. A simple approach might assert that two unfamiliar sounds are \emph{only} discriminable if they would be classified as distinct  phonemes by the listener. However, some unfamiliar sound contrasts are  easy. For example, English-speaking listeners have no difficulty with  Zulu click contrasts \cite{best1988examination} or the Thai [\textipa{W}]--[\textipa{7}] vowel contrast \cite{tyler2014perceptual}.

The Perceptual Assimilation Model (PAM: \citealt{best2007nonnative}) proposes that non-native speech perception is governed by the complete profile of listeners' assimilations to their native phoneme categories, which may be indeterminate. Thus, even if English has neither [\textipa{W}] nor [\textipa{7}], English-speaking listeners do not assimilate [\textipa{W}] to any specific English vowel, while they categorise [\textipa{7}] as English [\textipa{2}]. Since one of the two unfamiliar phones is not categorised as any native phoneme but the other is, listeners are able to discriminate them from one another.

PAM strongly implies that speech sound discrimination relies on phoneme-level representations cleansed of rich phonetic detail. However, an alternate approach proposes that listeners rely only on fine-grained, rich phonetic representations which are fine-tuned to the statistics of the native language, rather than mental categories \cite{guenther1996perceptual,iverson1995mapping,miller1997internal}. 
Recent modelling results by \citet{schatz2021early} have been interpreted as support for the view that, at least in early infancy, phoneme categories are not essential to explain effects of native language on speech perception. 

However, while there is evidence supporting each of these ideas, previous work has not directly compared  quantitative predictions of PAM against the predictions of a fine-grained phonetic account. 

We make use of the overlap score, an operationalization of PAM proposed by \citet{levy2009assimilation} which yields quantitative predictions of the degree of discriminability of an arbitrary pair of non-native phones from the results of an assimilation task performed on these phones. We then rely on recent developments in statistical machine learning of speech audio representations without the use of phoneme labels \cite{versteegh2016zero,dunbar2017zero}. We generate fine-grained statistically-driven representations of the experimental stimuli using a Dirichlet process Gaussian mixture model \cite{chen2015parallel}, the model proposed by \citet{schatz2021early} as an account of early phonetic learning. We also test wav2vec 2.0 \cite{Baevski2020wav2vec2A}, a more recent speech representation model. We compare these as predictors on a novel, open speech perception dataset containing French and English participants' behavioural results for 61 different vowels in Brazilian Portuguese, German, Turkish, Estonian, French and English. Our goal is to assess which model predicts participants' discriminability behaviour the best for the two groups, and, which one is able to predict the difference of behaviour between the two groups, presenting a native language effect.
 

Our results show that phoneme categories cannot be replaced by even the state of the art in fine-grained statistical phonetic modelling. While both the PAM-based predictor and the wav2vec 2.0 model are good predictors of the discriminability of stimuli, there is a small but consistent advantage for the assimilation-based predictor. Furthermore, the fine-grained models do not at all account for the differences in discriminability between the French and English listener groups, but, instead, seem to be improved predictors of a language-neutral ``phonetic similarity.'' The two factors, phonemic assimilation and phonetic similarity, are complementary, and both have a role to play in speech perception, consistent with other
evidence showing that listeners are sensitive to  fine-grained acoustic details at both neural and behavioural levels \cite{blumstein2005perception, pisoni1974reaction}.

The code and data we use are openly available.\footnote{\url{https://github.com/JAMJU/CONLL_2021_nonnative_speech_perception}}

\section{Discrimination dataset}
\label{sec:meth-discrimination}

To compare the predictions of PAM and fine-grained acoustic models, we create an experimental dataset that measures French- and English-speaking listeners' ability to discriminate native and non-native vowels. 
Participants judged stimuli constructed from twelve Brazilian Portuguese vowels 
([\textipa{\~i}], [i], [e], [\textipa{E}], [\textipa{\~e}], [\textipa{\~u}], [u], [\textipa{\~5}], [\textipa{\~o}], [o], [a], [\textipa{O}]), seven
standard German vowels 
([e], [i], [\textipa{E}], [a], [o], [\textipa{O}], [\textipa{a:}]), 
eight
Bavarian-accented (Munich) German vowels 
([\textipa{i:}], [\textipa{I}], [\textipa{Y}], [\textipa{U}], [\textipa{ø:}], [\textipa{y:}], [\textipa{u:}],[\textipa{a:}]), 
five
Turkish vowels 
([i], [y], [u], [\textipa{W}], [œ]),
six
Estonian vowels ([\textipa{i:}], [\textipa{ø:}], [\textipa{y:}], [\textipa{7:}], [\textipa{æ:}], [\textipa{a:}]), 
thirteen
French vowels 
([i], [\textipa{E}], [e], [œ], [a], [\textipa{\~O}], [y], [ø], [o], [u], [\textipa{O}], [\textipa{\~E}], [\textipa{\~A}]), 
and ten English vowels 
([i], [\textipa{I}], [\textipa{E}], [\textipa{eI}], [u], [\textipa{U}], [\textipa{2}], [\textipa{oU}], [æ], [\textipa{A}]).

These vowels were chosen to have a large coverage of the vowel space using a small number of languages, as well as to maximise the differences in perception behaviour between the French-speaking and the English-speaking listeners, as many front rounded and nasal vowels are included.

To test listeners' phone discrimination ability, we use an ABX discrimination task. The participant hears a stimulus A, a stimulus B, then a third one, X. They then have to decide which stimuli, A or B, is the closest to X, which always belongs to the same phone category as either A or B. Participants' accuracy reflects the distinctness of the A and B stimuli for the participants. In our case, French and English-speaking listeners' discrimination of pairs of native and non-native vowel sounds were tested, with A, B and X always belonging to the \textbf{same} language and using the same consonant frame (more details about the stimuli in Section \ref{sec:methods}). The results of the two groups were used to compare the relative predictive power of the PAM-type predictor versus the fine-grained model representations.

\section{PAM: From assimilation to discrimination}
\label{sec:methods}

According to the Perceptual Assimilation Model \cite{best2007nonnative} participants' behaviour during an assimilation task of non-native sounds can predict the discriminability of pairs of these same non-native sounds by listeners with the same language background. To derive such a predictor this, we conduct a separate experiment which elicits how the vowels listed in the previous section are assimilated by French- and English-speaking participants to their native vowel categories (see Section \ref{sec:meth-assimilation}). We apply the logic of PAM using the overlap score proposed by \citet{levy2009assimilation}: see Section \ref{sec:meth-overlap}.

\subsection{Phoneme assimilation}
\label{sec:meth-assimilation} 
According to PAM, naive listeners perceive unfamiliar sounds by mapping them (assimilating them) to their native phoneme category representations. In order to assess what phonemes listeners hear, an assimilation task is typically used---an experiment during which participants listen to a sound and must explicitly choose one of their native phonemes as a label. PAM asserts that pairs of unfamiliar sounds will be perceived as more or less distinct as a function of how they are assimilated to native phonemes.  During a discrimination task, participants should therefore discriminate pairs of non-native sounds well or poorly depending on the assimilation patterns associated with the two sounds.

We recruited French and English participants to perform an assimilation task using the same vowels as in Section \ref{sec:meth-discrimination}, but not exactly the same stimuli. For each phone, we calculate the percentage of participants who chose a given native phoneme in response (the phoneme to which they ``assimilated'' the given phone: only native vowels were possible responses). For each of the two listener groups, we obtain a \textbf{normalised vote vector} $v_p$ for each phone $p$ presented in the assimilation experiment, of dimension equal to the number of native vowel phonemes in the listener group's language. For a phone $p$ in the assimilation task and a native vowel phoneme $i$, $v_p[i]$ is the percentage of participants who assimilated $p$ as $i$ compared to all the other native vowel phonemes that were in the choice list. In particular, $\sum_i v_p[i] = 1$ .

\subsection{Predicting discrimination from phoneme assimilation patterns: Overlap score}
\label{sec:meth-overlap}

The overlap score is a method developed in \citet{levy2009assimilation} to quantitatively predict how difficult individual pairs of phones should be to discriminate for listeners, following PAM, based on the results of an assimilation task. It was created to generate predictions from PAM in a principled way without using arbitrary thresholds. See \citet{levy2009assimilation} for theoretical and empirical arguments in favour of the overlap score.

For a contrast $p_1$:$p_2$, one obtains the overlap score by first getting $v_{p_1}$ and $v_{p_2}$, the normalised vote vectors of the participants (see above for a definition). The overlap score can be computed as follows (with N the number of native phonemes used in the assimilation task):

\begin{align}
    \mathrm{overlap}(p_1, p_2) = \sum_{i = 1}^{N} \min(v_{p_1}[i], v_{p_2}[i])
\end{align}

For the contrast $p_1$:$p_2$, to be more consistent with the predicting values produced by acoustic models, we use the $\mathrm{neg-overlap}$ value as a predictor, which is defined as:
\begin{align}
    \mathrm{\text{neg-overlap}}(p_1,p_2) = - \mathrm{overlap}(p_1, p_2)
\end{align}
Thus, the bigger $\mathrm{\text{neg-overlap}}(p_1,p_2)$ is, the better humans are predicted to be at discriminating $p_1$ and $p_2$. The neg-overlap score obtained from French assimilation patterns is used to predict French discrimination patterns, and similarly for English. 

\section{Fine-grained phonetics: Predicting discrimination directly from audio}

We compare the experimentally derived neg-overlap score against several types of fine-grained phonetic representations from models (see Sections \ref{sec:dpgmm} and \ref{sec:wav2vec}) as predictors of listeners' discrimination behaviour. For each of the experimental stimuli, we calculate these representations for A, B, and X, and  compute a score called $\Delta$ (see Section \ref{sec:delta}), which measures how discriminable the representations are. They are considered more fine-grained than the neg-overlap score since they are more detailed in time (multiple representational frames per second compared to one assimilation judgement for a whole stimulus) and in dimensionality (around a thousand compared to around ten for the vote vectors used by the neg-overlap score).

\subsection{Raw acoustic features}
\label{sec:mfccs}
To give a baseline measure of discriminability based only on  acoustic similarity (not influenced by the statistics of any specific language), we use mel-frequency cepstrum coefficients (\textbf{MFCCs}). These representations are based on spectrograms on a mel scale (an approximation of the frequency scaling of the human auditory system). Like a spectrogram, they are sequences of continuous vectors (one MFCC vector every 10\,ms audio frame). We use the first 13 MFCC coefficients, calculated using \textsc{librosa},\footnote{https://librosa.org/} with a window of 25\,ms.

\subsection{Dirichlet process Gaussian mixture models}
\label{sec:dpgmm}

We use representations from a Dirichlet process Gaussian mixture model (\textbf{DPGMM}), a non-parametric Bayesian clustering model proposed as a speech representation model by \citet{chen2015parallel}. It finds, in an unsupervised way, a set of multi-dimensional Gaussian distributions appropriate to cluster (in our case) frames of acoustic feature vectors, taken at regular intervals (every 10\,ms). The model is trained on unlabelled speech recordings, so no phoneme categories are used. 

When applied to new speech recordings (in our case, the experimental stimuli), the model outputs a probability vector of dimension $K$ for each acoustic frame, where $K$ is the number of Gaussian distributions. Each element $k$ is the probability that the frame belongs to the $k^{\mathrm{th}}$ Gaussian distribution. Since there is one vector per acoustic feature frame, these representations are fine-grained in time. Furthermore, they are a rich representation (the models we use find more than 500 Gaussian distributions). DPGMM has been claimed to be a good model of early phonetic learning in infants \cite{schatz2021early}, and its representations have been shown to predict human listeners' behaviour in phone discrimination tasks: stimuli with more distinct DPGMM representations are generally easier for listeners to discriminate \cite{millet2019comparing,Millet2020PerceptimaticAH,Millet2020ThePE}.

We use two trained DPGMM models, one trained on English, the other on French. Both were trained on around 35 hours of speech, using MFCCs as input. We use the pretrained models provided by \citealt{millet2019comparing} (see paper), using the French model to predict French listeners' discrimination behaviour, and similarly for English.

\subsection{Wav2vec 2.0}
\label{sec:wav2vec}

Wav2vec 2.0 \cite{Baevski2020wav2vec2A} is a speech model that takes raw waveforms as input. It is composed of a convolutional feature encoder module, a quantization module using a learned codebook, and a context network made up of transformer layers. The model is learned in a self-supervised way, without any phoneme or textual labels, and, unlike the DPGMM, its representations integrate context, rather than treating audio frames as independent. See \citet{Baevski2020wav2vec2A} for details.

The model can also be fine-tuned to do phoneme recognition using labelled speech recordings. A supplementary linear layer is added at the end to predict the sequence of phoneme labels (for example, using connectionist temporal classification: \citealt{graves2006connectionist}). The resulting loss can be used to jointly update the phoneme recognition component and the previous layers of the wav2vec representation model \cite{wang2021voxpopuli,talnikar2021joint}.

We use three of the pretrained models described by \citet{wang2021voxpopuli} (base version). The first, \textbf{universal wav2vec}, is multilingual, trained in a self-supervised way on 23 languages, using ten thousand hours of unlabelled speech. We suppose that this model will give a better encoding of speech signals than the MFCC features, without being specific to a single language.

We also use two fine-tuned versions of this universal model, one trained on French (214 hours of transcribed speech recordings) and one on English (552 hours). We evaluate the French-trained model as a predictor of the French-speaking listeners' discrimination behaviour, and the English-trained model for the English-speaking listeners. We call this pair of models \textbf{native wav2vec}. While the representations obtained from these two models are influenced by phonemic information, they are still more fine-grained and less categorical than the assimilation patterns on which the overlap scores are based. They thus complement the DPGMM as an instantiation of the claim that rich phonetic representations, influenced by the statistics of the native language, can give rise to human-like native-language influences on speech perception.

We find that the representations at the fifth transformer block, which are temporally fine-grained sequences (one vector every 20\,ms) of rich (768-dimensional) vectors, give the best predictions on our discrimination experiment, for all three models. We  retain these representations for the remainder of the paper.

\subsection{Predicting discrimination from fine-grained acoustic modelling: $\Delta$ score}
\label{sec:delta}

To predict discrimination ability for a triplet \emph{target}--\emph{other}--X, where \emph{target} and X are examples of the same vowel phone category, and \emph{other} is from a different category, we compute the distance between a model's representation of \emph{target} and X, and the distance between its representation of \emph{other} and X, using a distance function $D$. We define $\Delta$ as the subtraction of these two distances.
\begin{align}
   \Delta = D(other,X) - D(target,X)
\end{align}
The bigger $\Delta$ is, the better listeners are predicted to be at accurately performing the ABX discrimination for the given experimental item.

The representations of the stimuli obtained from the models we use are time sequences which do not all have the same length. We calculate $D$ using dynamic time warping based on the symmetrised KL-divergence (\textbf{DPGMM}) or the cosine distance (\textbf{MFCCs}, \textbf{wav2vec}), as in previous works \cite{millet2019comparing,Millet2020PerceptimaticAH,Millet2020ThePE}.

\section{Methods: Discrimination and assimilation experiments}
\label{sec:methods}
In this section we provide basic information about the experiments with human participants. For more details, see the Appendix. 

\subsection{Stimuli}

Native speakers of English, French, Brazilian Portuguese, Turkish, Estonian and Bavarian-accented German recorded consonant-vowel-consonant (CVC) stimuli in carrier sentences. We also used recordings of six speakers taken from the OLLO database \cite{ollo}, in standard German. Together, we use these CVC stimuli as the basis for items in both a discrimination and an assimilation experiment, described below. Some phones from different languages shared the same IPA symbol (for example, Turkish and French [i]). We treat these as distinct phones for the purposes of our analyses. All the stimuli were resampled to 16000\,Hz, and we equalised their volume.

\subsection{Discrimination task}
\label{sec:exp-discrim}

The participants performed an ABX test (see Section \ref{sec:meth-discrimination}). For each trial, the three stimuli A, B, and X were CVCs which differed only in the central vowel, and had the same consonant frame. The target response (either A or B) matched X in its central vowel, while the other response was a different vowel phone. A, B, and X were always taken from the same language (the two German stimuli sets were treated as separate). A and B were uttered by the same speaker, and X by a different speaker. A 500\,ms silence separated A and B, and 750\,ms separated B and X. We added speech-shaped noise to obtain a signal-to-noise ratio of 5\,dB in order to avoid a ceiling effect. 

We tested twenty contrasts from Brazilian Portuguese, five from standard German, twelve from Bavarian-accented German, eight from Turkish, eleven from Estonian, thirteen from French, and ten from English, for a total of 79 contrasts. Each contrast was presented to the participants in three different consonant frames, and using three different combinations of speakers. We presented the stimuli both in the order \emph{target}--\emph{other}--X (with the correct answer being A) and in the order \emph{other}--\emph{target}--X (with the correct answer being B). Each combination was judged by five participants. Thus, in total, each contrast was tested at least 180 times.

\subsubsection{Assimilation task}

\label{sec:exp-assim}

As alluded to above, in order to construct the empirically-derived neg-overlap scores, we also tested participants on the way they assimilated different vowel sounds to their native vowel phonemes, in a separate experiment.
In total, we tested 61 vowels (see Section \ref{sec:methods}). Each vowel was included in the assimilation test with four different contexts (consonant frames) and four different speakers (16 different stimuli for each vowel), except for the German [\textipa{I}], for which only three contexts were recorded, but for which six speakers' recordings were used, resulting in 18 stimulus tokens. We used the same stimuli as in the discrimination task as much as we could. Each individual stimulus was tested with five different participants for each language (French and English), counterbalanced so that each vowel was tested at least $5\times16 = 80$ times per language. For each CVC stimulus they heard, participants had to indicate, using reference words, which of their native vowels the central vowel of the stimulus was the closest to (see the appendix for the detailed list of words used).

\subsection{Participants}

We recruited French- and English-speaking participants to perform each of the two tasks. The two tasks were not done sequentially, and were done independently, by different sets of participants. Both experiments were done online.

We tested English-speaking participants from the United States and French-speaking participants from France. We offered to pay each participant that finished the study (some participants declined). We recruited participants through Amazon Mechanical Turk (MTurk), social networks and in person. We rejected participants who were not monolingual native speakers of the language tested, who were fully bilingual in another language (considering themselves as native speakers of the other language), as well as participants who failed too many catch trials. After this filtering, for the discrimination task, we were left with 81 French-speaking participants (48 from Mturk, 33 other) and 80 English-speaking participants. For the assimilation task, we were left with 68 participants (50 from MTurk, 18 in person) for French, and 70 participants for English.

\section{Results}

\label{sec:eval}

We compare the neg-overlap score with $\Delta$ scores from fine-grained phonetic models. We seek to assess how well they can predict the results of the discrimination experiment. This comparison is done using two metrics: the \textbf{log-likelihood} of a binary regression model applied to the results of the experiment, and the \textbf{Spearman's} $\rho$ \textbf{correlation} between the average of the given predictor and listeners' accuracy for each pair of phones. 

We calculate the log-likelihood ($\ell\ell$) using a probit model, fitted to the complete set of discrimination responses for each listener group, with correct responses coded as 1 and incorrect responses as 0. We fit one model for each of the neg-overlap and $\Delta$ scores we seek to compare, with the given score as a predictor: each observation is paired with the score for the stimulus heard by the participant. In the case of neg-overlap, the score is calculated for the pair of phones using the results of the assimilation experiment, while the $\Delta$ scores are calculated on the A, B, and X audio files. In addition to a global intercept, the model has additional predictors accounting for effects of a number of nuisance factors: whether the right answer was A (1) or B (0); the position of the trial in the experimental list; and a categorical predictor for the participant. Higher (less negative) log-likelihood values indicate better predictions.

We complement the log-likelihood with a correlation statistic, which has the advantage of being on the interval between 0 and 100\%, and having been used in previous works \citep{levy2009assimilation}. Consistent with \citet{levy2009assimilation}, we calculate the correlation at the level of phone pairs, rather than individual items,\footnote{We carried out a similar analysis using a probit model by substituting the average $\Delta$ score for the item-level $\Delta$ score. The results were very similar to the item-level analysis presented here.} and we use Spearman's $\rho$, a correlation between ranks: in this case, the rank of the discrimination accuracy, versus the rank of the neg-overlap or average $\Delta$ score, for the given phone pair.

To assess the statistical significance of the differences between model comparison scores, we perform a bootstrap resampling of the participants' results: for each item used in the discrimination experiment, we resample the results with replacement in order to obtain five scores. We generate $N$ bootstrap samples. For each sample, we calculate $\ell\ell$ ($\rho$) for all predictor scores, and then compute the difference in $\ell\ell$ ($\rho$) to compare two predictors. If the 95 \% bootstrap confidence interval of the differences does not contain zero, we take the difference to be significant.

\subsection{Human discrimination patterns}

\begin{figure}[h!]
    \centering
    \includegraphics[trim ={0 0 2.1cm 3cm}, clip, scale = 0.1 ]{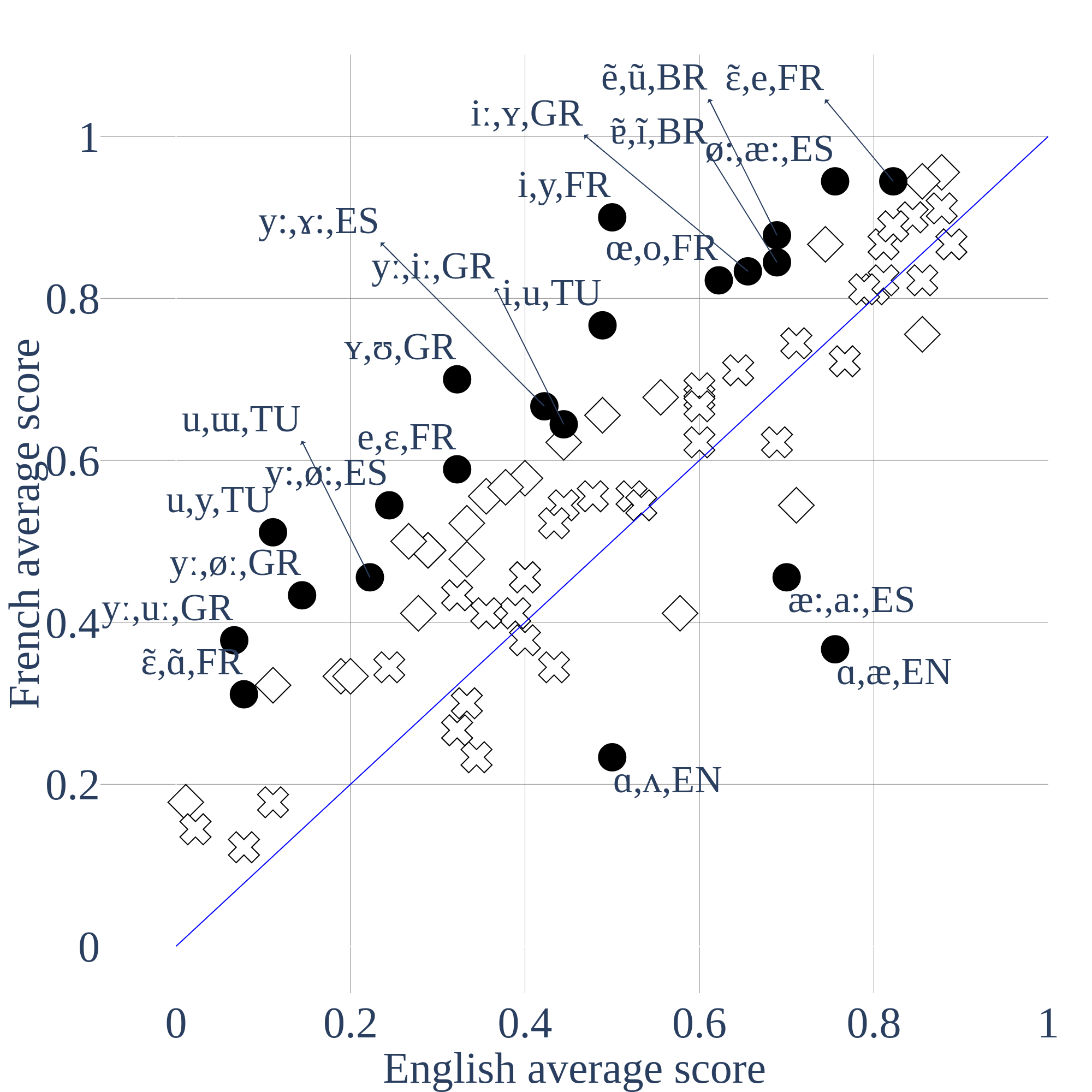}
    \caption{French discrimination results versus English discrimination results. Zero represents chance-level, for each item, as participants receive a score of 1 if they are correct, -1 otherwise. The line represents $y = x$. Each point is a phone pair, the median of a bootstrap resample ($N$=10000). If a pair is above the line, then French native listeners are better at discriminating it; otherwise, English native listeners are better. Black circles are contrasts that are significantly different between the two groups. White diamonds are contrasts that are significantly above or below $y=x$; white crosses are contrasts that are not significantly different for the two groups.}
    \label{fig:fr_vs_en}
\end{figure}

Discrimination results for each contrast studied, for French against English participants, can be seen in Figure \ref{fig:fr_vs_en}. The results obtained are consistent with previous studies, with English participants having trouble with the contrasts [i]/[y] and [y]/[u] in different languages \cite{levy2008perception,tyler2014perceptual}, and French participants with the contrasts [\textipa{2}]/[\textipa{A}], [æ]/[\textipa{A}] and [\textipa{æ:}]/[\textipa{a:}] \cite{iverson2012auditory,millet2019comparing}. Other contrasts significantly different for the two groups can be seen in the figure (black circles).

\subsection{Comparing PAM with fine-grained phonetics}

\label{sec:human_behaviour}
\begin{figure}
   \centering
    \vskip -1cm
    \includegraphics[trim ={0.2cm 0cm 0cm 1cm}, clip,scale = 0.51]{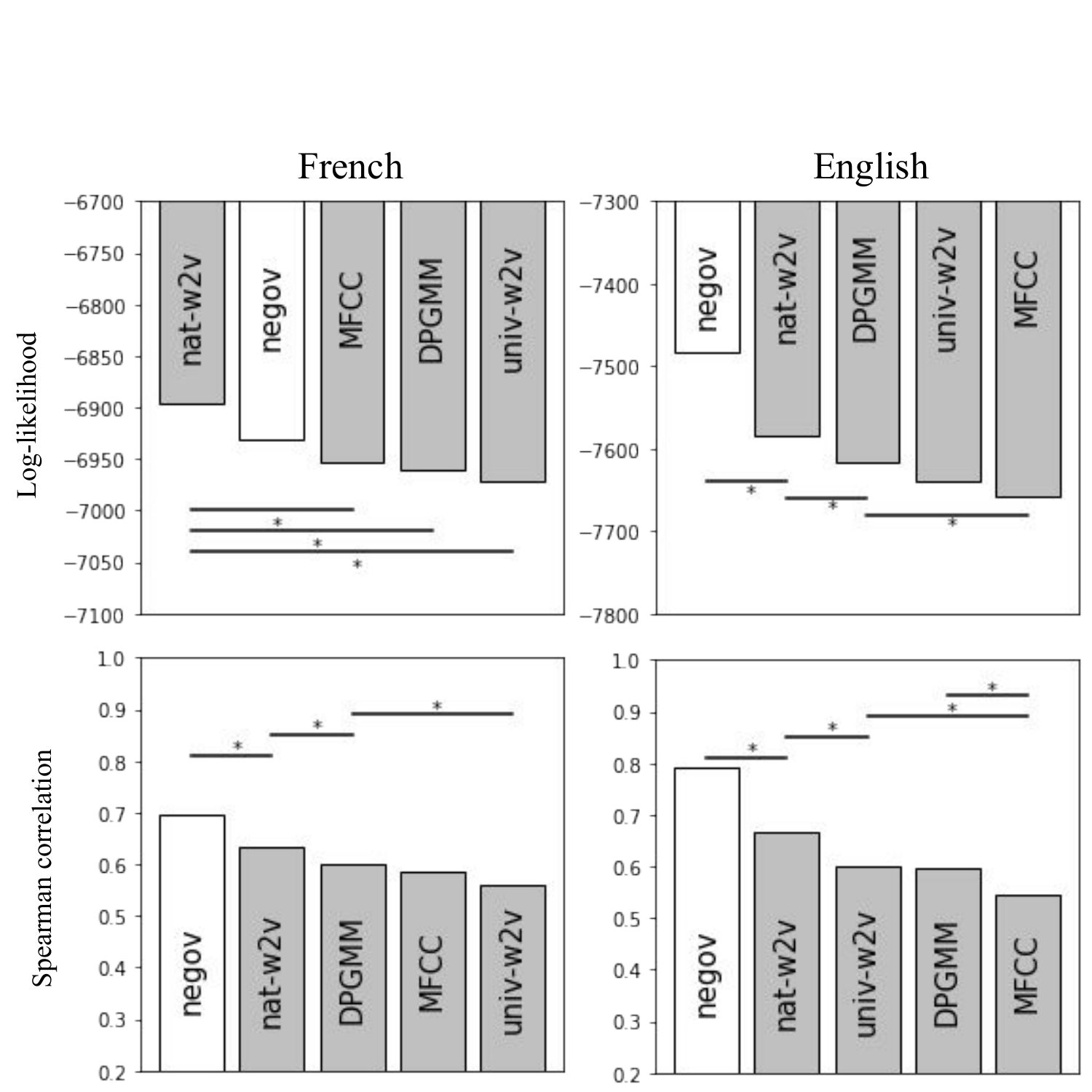}
    \caption{Log-likelihood values (top: shorter bars are better) and Spearman correlation (bottom: taller bars are better) for French (\emph{left}) and English participants (\emph{right}). Stars indicate that the pairwise difference is significant. Redundant statistical comparisons are omitted for clarity (i.e. $C>A$ is omitted when $C>B$ and $B>A$). The assimilation-based neg-overlap score is in white to distinguish it from the fine-grained phonetic representations, in grey.}
    \label{fig:results}
\end{figure}

We compare the neg-overlap obtained from the assimilation patterns with the $\Delta$ values computed from fine-grained phonetic representations. The log-likelihood values obtained by the fitted probit models and the Spearman correlations can be seen in Figure \ref{fig:results}. The neg-overlap score is statistically significantly better than all the phonetic predictors in most cases, except for the comparison by log-likelihood for the French participants. In general, however, the neg-overlap score gives good results for both language groups compared to the fine-grained models' $\Delta$ values, supporting the idea of assimilation to phoneme categories. We  note, however that, since the neg-overlap score is obtained using human participants' assimilation patterns, its predictive performance should have been close to perfect if assimilation to phoneme categories were the only mechanism driving discrimination behaviour. This is not the case: see Section \ref{sec:combining}.

\textbf{Native wav2vec} is statistically significantly better than the other fine-grained phonetic representations, by both measures, for both groups. There is no clear winner among the other three predictors. For French, none of the remaining $\ell\ell$ value differences are significant, and the only significant $\rho$ difference is \textbf{DPGMM} versus \textbf{universal wav2vec} (\textbf{DPGMM} is better). For English, only the $\ell\ell$ advantage for \textbf{universal wav2vec} over \textbf{MFCC} is significant, and the advantages in $\rho$ for both the \textbf{DPGMM} and \textbf{universal wav2vec} over \textbf{MFCC}.

\begin{figure*}[ht]
    \centering
    \includegraphics[scale = 0.15]{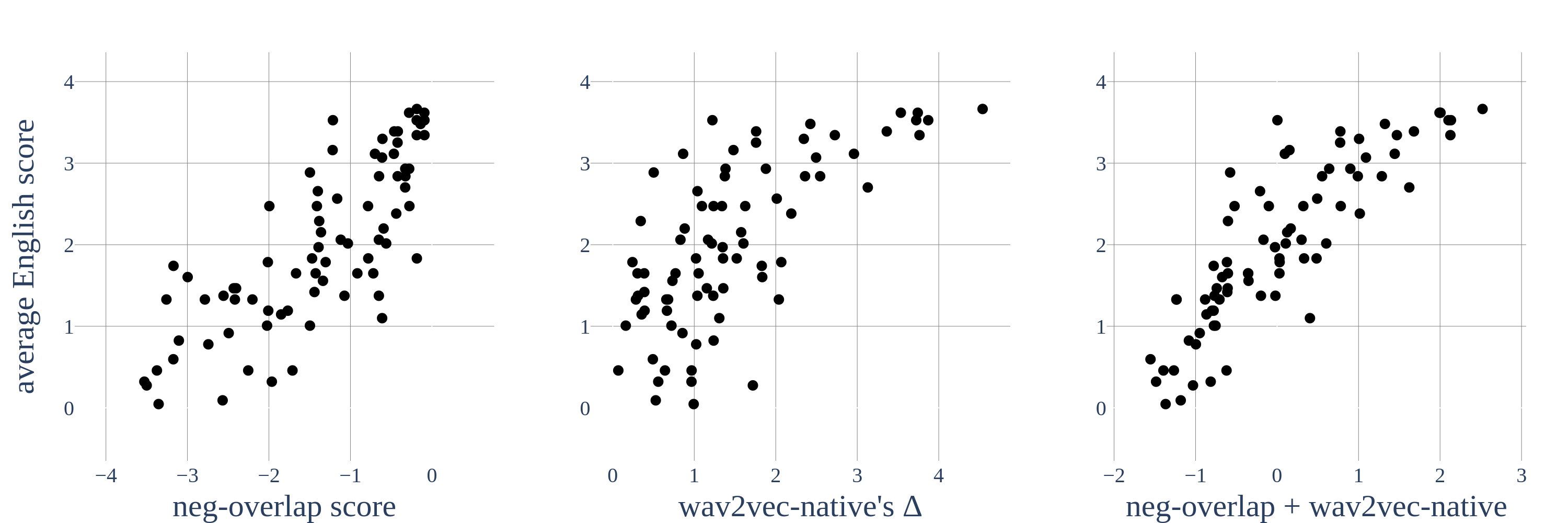}
    \caption{English participants' average result per contrast (0 is chance level, the bigger the better participants are at discriminating the considered contrast), versus the average neg-overlap score (\emph{left}), $\Delta$ values from \textbf{native wav2vec} (\emph{middle}) and the sum of the two after normalisation (\emph{right}). Values are normalised (divided by the standard deviation) to be on the same scale. Human listeners' values are the medians obtained by bootstrapping ($N=10000$).}
    \label{fig:three_graph_correl}
\end{figure*}

Neg-overlap values are plotted against English participants' results, averaged for each contrast at left in Figure \ref{fig:three_graph_correl}, and the comparison against the $\Delta$ values from \textbf{native wav2vec} is shown in the middle.

\subsection{Capturing native language effects}
\label{sec:native-effect}

Assimilation patterns and \textbf{native wav2vec} both seem to predict
participants' phone discrimination well. However, we seek to assess whether they also capture the differences between the two listener groups. Importantly, while both outperform the simple acoustic distances calculated on the MFCC representations in predicting discriminability, they may nevertheless be principally explaining behaviour shared by both listener groups, driven by universal phonetics. This is possible for the neg-overlap score because much of the assimilation behaviour is comparable for French and English listeners. It is also possible for \textbf{native wav2vec,} because the effects of fine-tuning on a specific language need not necessarily have negative impacts for other languages, and may indeed be beneficial. 

We thus evaluate which of the scores best predicts the effect of native language: the \emph{differences} between the two groups. In other words, when testing the native effect captured by \textbf{native wav2vec}, we want to see if for a particular contrast that is difficult for French speaking participants but easy for English speaking participants, the English wav2vec's $\Delta$ value is bigger (`easier' for the model), in proportion, than the French wav2vec's $\Delta$ value (`harder' for the model).

To test this systematically, we calculate, for each phone contrast, the subtraction of the average French-based and English-based predictor scores, for each of the three language-specific predictors. (We divide the predictor's values by its standard deviation to be comparable across languages.) To obtain the difference of discrimination behaviour between the two listener groups, we calculate the subtraction of the percent accuracy between the French- and the English-speaking participants. We then calculate the Pearson correlation $r$ between the differences calculated for models and for humans. Correlations and 95\% confidence intervals can be seen in Table \ref{tab:native-effect}.

\begin{table}[h!]
\centering
\begin{tabular}{lcc}
\hline
\textbf{Model} & Median $r$ & 95\% CI\\
\hline
Neg-overlap score & 0.66 & $[0.58,0.71]$\\
Native wav2vec & 0.05 & $[0.002,0.12]$\\
DPGMM & 0.05 & $[0.002,0.11]$ \\ \hline
\end{tabular}
\caption{Native language effect evaluation: Pearson correlation between the differences in French versus English behaviour, for the two listener groups, and for the trained models and neg-overlap scores. The bigger $r$, the better the model/value encodes native language specificities, and predicts specific native listeners' discrimination behaviour. 95\% intervals are obtained using a bootstrap with $N=10000$.}
\label{tab:native-effect}
\end{table}

 The correlation is clear for the neg-overlap score, but it is almost nonexistent for the \textbf{native wav2vec} and \textbf{DPGMM} models: the fine-grained phonetic models capture almost none of the native language effects. For \textbf{DPGMM,} the result is a surprise, since previous studies found native language effects \cite{schatz2021early,millet2019comparing}. However, the item-level effect found in \citet{millet2019comparing} was very small, and disappeared when stimuli were grouped by phone contrast. \citet{schatz2021early} also found very small effects of native language using \textbf{DPGMM}, as compared to a speech recogniser trained in a supervised way to predict phoneme labels. The result for \textbf{native wav2vec} is therefore surprising, since that model is fine-tuned using phoneme labels. However, as mentioned above, there is no guarantee that a model's specialization for one language's phoneme categories will imply a loss of sensitivity to another's; the degree to which this happens, and to which it matches human behaviour, may differ as a function of the phoneme contrast and the stimuli used. See also \citet{Millet2020PerceptimaticAH}.

\subsection{Combining phonology and phonetics}
\label{sec:combining}

The fine-grained phonetic models do not seem to capture the differences in behaviour between the English- and French-speaking participants. Yet, particularly for \textbf{native wav2vec}, they do seem to make good predictors of discrimination patterns. This means that \textbf{native wav2vec's} perceptual space does a good job at modelling the universal component of human speech perception deployed when listeners perform a discrimination task---better than the generic \textbf{MFCC} acoustic representations.

Some models of speech perception propose that discrimination tasks simultaneously tap into phonemic and universal acoustic/phonetic representations (for example, \citealt{fujisaki1970some,pisoni1974reaction}). This would imply that the neg-overlap and fine-grained scores are complementary. Alternatively, it is possible that the influence of phonetics is already completely captured by the assimilation profiles, which are probabilistic, and thus capture gradient information. In this case, the fine-grained phonetic models would just be (poor) approximations of the neg-overlap score.

To assess whether the information provided by the phonetic models is already entirely present in the neg-overlap score, we use both the neg-overlap score and the native models' $\Delta$ values to predict human behaviour at the stimulus level, using them both as predictors in a probit model. We assess how much the log-likelihood obtained improves compared to using either the overlap score or the \textbf{native wav2vec} $\Delta$ values alone. The $\ell\ell$ improves statistically significantly over both single-predictor models for both English-speaking ($-7347.5$, from $-7483.9$ and $-7584.0$) and French-speaking ($-6759.1$, from $-6931.7$ and $-6896.0$) participants. We thus conclude that the two are complementary, and capture different aspects of speech discrimination. This can be seen qualitatively in the rightmost plot in Figure \ref{fig:three_graph_correl}, which plots English participants' average performance by contrast against the sum of the overlap score and \textbf{native wav2vec} $\Delta$ values: the correlation is visibly improved over either of the two predictors individually.

\section{Discussion and summary}

We have presented a dataset of non-native speech perception combining a native phoneme assimilation task with a discrimination task, appropriate for exploring questions about the role of native language categories in speech perception. The data is open, and contains a large number of vowels, presented to French- and English-speaking listeners. 

Our analysis shows that French and English speakers' patterns of assimilation to native vowel phonemes are good predictors of discriminability by naive listeners, consistent with the Perceptual Assimilation Model. The prediction, however, is not perfect.

We then did a comparable analysis using rich, fine-grained phonetic models obtained by applying machine learning to speech, which do not directly use coarse-grained phoneme categories. 

We showed that none of these models is consistently better at predicting listeners' discrimination behaviour than the experimentally-derived assimilation scores, and, contrary to previous studies, they show only minimal effects of the ``native'' language on which they were trained. The assimilation scores, in contrast, are good at capturing the differences between our French- and English-speaking listeners.

Finally, we demonstrate that the \textbf{native wav2vec} model is a good model of universal phonetics, and provides information about phonetic similarity which is complementary to the phoneme assimilation-based scores, consistent with the very old idea that discrimination experiments are explained by a mix of categorical effects and phonetics \cite{fujisaki1970some}. Compared to previous attempts to measure phonetic similarity, the wav2vec model does not require choosing specific phonetic measures in advance (like formants), and is applicable to any speech stimuli.

\section{Acknowledgements}
This research was supported by the \'Ecole Doctorale Frontières du Vivant (FdV) -- Programme Bettencourt, and by grants ANR-17-CE28-0009 (GEOMPHON), ANR-11-IDFI-023 (IIFR), ANR-11-IDEX-0005 (USPC),  ANR-10-LABX-0083 (EFL),  ANR-17-EURE-0017 Frontcog, ANR-10-IDEX-0001-02 PSL*, ANR-19-P3IA-0001 PRAIRIE 3IA Institute. This work was performed using HPC resources from GENCI-IDRIS (Grant 20XX-AD011012415). 
\bibliography{anthology,custom}
\bibliographystyle{acl_natbib}
\clearpage
\newpage
\appendix

\section{Appendix}
\label{sec:appendix}
\begin{table*}[]
\centering
\begin{tabular}{|l|c|c|c|}
\hline
Language & Carrier sentence & Triphone's transcription & Volunteers \\ \hline
American English & I like {[}...{]} here & orth. for reference words, IPA for the rest & 4F, 4M \\ \hline
Brazilian Portuguese & Eu digo {[}...{]} aqui & IPA precised, orth. in sentences & 2F, 3M \\ \hline
Estonian & Ma ütlen {[}...{]} siin & orth. & 3F, 2M \\ \hline
German (Munich) & Ich sage {[}...{]} hier & IPA & 3F, 3M \\ \hline
Metropolitan France's French & Je dis {[}...{]} ici & orth., except for œ, ø, o and \textipa{O} & 4F, 4M \\ \hline
\end{tabular}
\caption{Details of the career sentences used for each language, the type of transcriptions used by the volunteers (orth is orthographic) and the number and sex of the volunteers (F: female, M: male).}
\label{tab:career}
\end{table*}

\begin{table}[]
\centering
\begin{tabular}{|c|c|}
\hline
Vowel & Reference word \\ \hline
/i/ & t\textbf{i}que \\ \hline
/y/ & t\textbf{u}t\textbf{u} \\ \hline
/u/ & t\textbf{ou}j{ou}rs \\ \hline
/e/ & journ\textbf{é}e \\ \hline
/\textipa{E}/ & cach\textbf{e}tte \\ \hline
/ø/ & m\textbf{eu}nier, qu\textbf{eu}e \\ \hline
/œ/ & v\textbf{eu}ve et s\textbf{oeu}r \\ \hline
/a/ & p\textbf{a}p\textbf{a} \\ \hline
/o/ & rid\textbf{eau} \\ \hline
/\textipa{O}/ & p\textbf{o}rte, s\textbf{o}rt \\ \hline
/\textipa{\~E}/ & \textbf{im}pair \\ \hline
/\textipa{\~O}/ & pard\textbf{on} \\ \hline
/\textipa{\~A}/ & méch\textbf{an}t \\ \hline
\end{tabular}
\caption{Reference words used for French speaking participants}
\label{tab:ref_french}
\end{table}

\begin{table}[]
\centering
\begin{tabular}{|c|c|}
\hline
Vowel & Reference word \\ \hline
/\textipa{I}/ & k\textbf{i}t \\ \hline
/\textipa{E}/ & dr\textbf{e}ss \\ \hline
/æ/ & tr\textbf{a}p \\ \hline
/\textipa{A}/ & st\textbf{o}p \\ \hline
/\textipa{2}/ & str\textbf{u}t \\ \hline
/\textipa{U}/ & l\textbf{oo}k \\ \hline
/i/ & fl\textbf{ee}ce \\ \hline
/u/ & l\textbf{oo}p \\ \hline
/o\textipa{U}/ & g\textbf{oa}t \\ \hline
/\textipa{eI}/ & t\textbf{a}pe \\ \hline
\end{tabular}
\caption{Reference words used for English speaking participants}
\label{tab:ref_english}
\end{table}

\section{Detailed methods: Discrimination and assimilation experiments}

\subsection{Stimuli}
We asked volunteers, native speakers of either English, French, Brazilian Portuguese, Turkish, Estonian or German (speakers from Munich), to record a list of stimuli in carrier sentences (details in the appendix). The same stimuli are used in both experiments and are of the form consonant-vowel-consonant (CVC). Volunteers recorded the stimuli multilet times on their own computers, using their own microphones, in a quiet environment of their choice.

We used the Montreal Forced Aligner \cite{montrealforced} to align the recordings with the read text for each volunteer, resulting in a phone-level annotation. We then cut out each CVC, and we checked the quality of the recordings and the alignment by hand for each stimulus. 

We also used recordings taken from the OLLO database \cite{ollo}, in German. The stimuli were already cut out. We used six speakers who speak standard German: S01F, SO4F, S07F, S02M, SO3M, S05M (three female, three male). We used the normal speaking rate version of their recordings.

All the stimuli were resampled to 16000\,Hz, and we equalised their volume.

\subsection{Discrimination task}
\label{ap:exp-discrim}
 The participants had to perform an ABX test (see Section \ref{sec:meth-discrimination}). Instead of simply answering A or B, they had to answer on a six-point scale ranging from `first stimuli for sure' to `second stimuli for sure.' We considered their response to be 1 if they were correct, and -1 if they were wrong.

 For each trial, the three stimuli A, B, and X are CVCs which differ only in the central vowel, and have the same consonant frame. The target response (either A or B) matches X in its central vowel, while the other response is a different vowel phone. All three phones are taken from the same language (the two German stimuli sets are treated as separate). A and B are uttered by the same speaker, and X by a different speaker. A 500\,ms silence separates A and B, and 750\,ms separates B and X. We add speech-shaped noise to obtain a signal to noise ratio of 5\,dB. 
 
 We tested twenty contrasts from Brazilian, five from standard German, twelve from Bavarian-accented German, eight from Turkish, eleven from Estonian, thirteen from French, and ten from English. In total we tested 79 contrasts. Each contrast was presented to the participants in three different consonant frames,\footnote{The complete list of contexts used for each contrast can be found on our Github.} and using three different combinations of speakers. We presented the stimuli both in the order target--other--x (with the correct answer as A) and in the order other--target--x. Each combination was judged by five participants. Thus, in total, each contrast was tested at least 180 times.\\
 
 No participant was tested twice on the same phone pair in the same context. The same combinations of speakers recur across multiple trials, but the experimental list for each participant is constructed so that the combination of speakers was not predictive of the right answer.
 
Before the experiment, participants were trained using simple stimuli (`cat' and `dog' for English, `hibou' and `caillou' for French) and native-language CVC stimuli, without noise, followed by feedback. Then, they hear an example of a non-native item with noise. 

During the experiment, participants were tested on 158 to 180 items. This number differed because of counterbalancing. Every 39 items, an easy catch trial was included (using the same stimuli as in the training). The same thing was done every 17 items for catch trials using easy native contrasts, without noise. We used these to filter out participants, see Section \ref{ap:recruiting}

\subsubsection{Assimilation task}
\label{ap:exp-assim}
In total we tested 61 vowels (see Section \ref{sec:methods}). Each vowel was included in the assimilation test with four different contexts (consonant frames) and four different speakers (16 different stimuli for each vowel), except for the German \textipa{I} for which only three contexts were recorded but for which six speakers' recordings were used, resulting in 18 stimulus tokens.\footnote{The complete list of contexts used for each vowel can be found on our Github.} Each individual stimulus was tested with five different participants for each language (French and English). Thus, each vowel was tested at least $5\times16 = 80$ times per language.

We tested participants on the way they assimilated different vowel sounds to their native vowel phonemes.
For each CVC stimulus they heard, they had to vote, using reference words (see the appendix for a list of the used French and English reference words), to indicate which of their native vowels the central vowel of the heard stimulus was the closest to. Then, on a different page, they had to score the stimulus heard, assessing how good it was as a typical example of the vowel sound they selected, rating it from 1 (very unusual) to 5 (typical example). Here, we only use the votes, ie the one-hot choice each participant made, and not the scores.

Before doing the experiment, participants first had to listen to reference words for each of their native vowels, in order to be able to recognise them in the stimuli we gave them afterwards. Then, they trained for the task, first judging each of the vowels from their native language (ten for English participants, thirteen for French), following the process described above, then one non-native example. After this training, the experiment started. They performed the assimilation task on 63 to 75 stimuli. This number differed because of counterbalancing. Three breaks were included in the experiment, during which participants could rest. During the breaks, they also had to listen to the reference words for each vowel again, uttered by another speaker. There were five catch trials during the test in order to test participants' attention, and filter them out, see Section \ref{ap:recruiting}. For each of them, participants had to listen to either `caillou' or `hibou' for French, or `cat' or `dog' for English, and choose among two words which one they heard.

\subsection{Participants}
\label{ap:recruiting}
We tested English-speaking participants from the United States and French-speaking participants from France. We offered to pay each participant that finished the study (some participants declined). We recruited participants through Amazon Mechanical Turk (MTurk), social networks and in person. 

The participants answered a presurvey on their language background. This allowed us to reject participants who were not native speakers of the language tested (French or English), who were fully bilingual in another language (considering themselves as native speakers of the other language), or who were intensively exposed to another language as a child (before eight years old). For the discrimination task, we also rejected participants who made more than one mistake on the easy catch trials or made more than one mistake on the easy native catch trials. For the assimilation task, we also rejected participants if they made more than one mistake on the catch trials or if they correctly identified less than half of the items that were drawn from their native language (either English or French) during the test (the experiment contained either 15, 16 or 17 such stimuli for French and either 11, 12 or 13 for English).

After this filtering, for the assimilation task, we were left with 68 participants (50 from MTurk, 18 in person) for French, and 70 participants for English. For the discrimination task, we were left with 81 French-speaking participants (48 from Mturk, 33 other) and 80 English-speaking participants.

\end{document}